%% file: SemDI.tex
\pdfoutput=1

\documentclass[11pt]{article}

\usepackage{acl}

\usepackage{times}
\usepackage{latexsym}

\usepackage[T1]{fontenc}

\usepackage[utf8]{inputenc}

\usepackage{microtype}

\usepackage{inconsolata}

\usepackage{amsmath}
\usepackage{booktabs}
\usepackage{multirow}
\usepackage{graphicx}
\usepackage{enumerate}
\usepackage{subfigure}
\usepackage{threeparttable}
\usepackage{amssymb}
\usepackage{amsmath}
\usepackage{mathrsfs}
\usepackage{float}
\usepackage{algpseudocode}
\usepackage{balance}
\usepackage{enumerate}
\usepackage{enumitem}
\usepackage{footnote}
\usepackage{xcolor}
\usepackage{makecell}
\usepackage{dsfont} 
\usepackage{bbding}
\usepackage{tabularx}
\usepackage{stfloats}

\title{Advancing Event Causality Identification via Heuristic Semantic Dependency Inquiry Network}

\author{Haoran Li\textsuperscript{1}, Qiang Gao\textsuperscript{2,3}, Hongmei Wu\textsuperscript{2}, Li Huang\textsuperscript{2,3}\thanks{~~Corresponding author (lihuang@swufe.edu.cn).} \\
       \textsuperscript{1}College of Computer Science, Sichuan University
       \\
       \textsuperscript{2}School of Computing and Artificial Intelligence, \\Southwestern University of Finance and Economics\\
       \textsuperscript{3}Engineering Research Center of Intelligent Finance, Ministry of Education, \\
       Southwestern University of Finance and Economics\\
       \texttt{haoran.li.cs@gmail.com},
       \texttt{\{qianggao, lihuang\}@swufe.edu.cn}}
        
\begin{document}
\maketitle
\begin{abstract} 
Event Causality Identification (ECI) focuses on extracting causal relations between events in texts. Existing methods for ECI primarily rely on causal features and external knowledge. However, these approaches fall short in two dimensions: (1) causal features between events in a text often lack explicit clues, and (2) external knowledge may introduce bias, while specific problems require tailored analyses. To address these issues, we propose SemDI - a simple and effective {\bf Sem}antic {\bf D}ependency {\bf I}nquiry Network for ECI. SemDI captures semantic dependencies within the context using a unified encoder. Then, it utilizes a {\it Cloze} Analyzer to generate a fill-in token based on comprehensive context understanding. Finally, this fill-in token is used to inquire about the causal relation between two events. Extensive experiments demonstrate the effectiveness of SemDI, surpassing state-of-the-art methods on three widely used benchmarks. Code is available at \url{https://github.com/hrlics/SemDI}.

\end{abstract}

\section{Introduction}
\label{introduction}
\input{1-Introduction}

\section{Related Work}
\label{related_work}
\input{2-Related_work}

\section{Preliminaries}
\label{Preliminaries}
\input{3-Preliminaries}

\section{Methodology}
\label{Methodology}
\input{4-Method}

\section{Experiments}
\label{Experiments}
\input{5-Experiments}

\section{Conclusions}
In this paper, we present SemDI, a simple and effective semantic dependency inquiry approach for Event Causality Identification. We first encode the semantic dependencies using a unified encoder. Subsequently, we utilize a {\it Cloze} Analyzer to generate a fill-in token based on comprehensive context understanding. This token is then used to inquire about the causal relation between two events within the context. Extensive experiments on three widely recognized datasets demonstrate the superior performance of SemDI while highlighting its robustness and efficiency.

\section*{Limitations}
The limitations of this work can be concluded as follows:
\begin{enumerate}
    \item SemDI exhibits sensitivity to the quantity of annotated event pairs available for training. Consequently, it demonstrates reduced accuracy in capturing causal relations within the CTB dataset, as illustrated in Table.~\ref{tab:res-ctb}. Therefore, further improvements are needed to enhance its performance in low-resource scenarios.
    \item While acknowledging the potential for bias introduced by external knowledge, we argue that incorporating commonsense is crucial for ECI. SemDI concentrates on investigating the effectiveness of semantic dependency inquiry for ECI, leaving the opportunity to take advantage of commonsense reasoning. Investigating how to properly integrate commonsense reasoning within the semantic-guided framework presents a promising avenue for future research.
\end{enumerate}


\section*{Acknowledgements}
This work was supported by the Guanghua Talent Project.

\bibliography{anthology, latex/custom}

\clearpage
\appendix
\section{Appendix}
\label{app}

\input{6-Appendix}

\end{document}

%% file: 1-Introduction.tex
Event Causality Identification (ECI) aims to catch causal relations between event pairs in text. This task is critical for Natural Language Understanding~(NLU) and exhibits various application values. For example, an accurate ECI system can facilitate question answering \citep{liu2023cross, zang2023discovering}, narrative generation \citep{ammanabrolu2021automated}, and summarization \citep{huang2023causalainer}. However, identifying causal relationships within text is challenging due to the intricate and often implicit causal clues embedded in the context.  For instance, in the sentence "{\it But \textbf{tremors} are likely in the junk-bond market, which has helped to finance the takeover \textbf{boom} of recent years.}", an ECI model should identify the causal relation between event pair {\it \textbf{(tremors, boom)}}, which is not immediately evident without understanding the context.

\begin{figure}[t!]
    \centering    \includegraphics[width=0.45\textwidth]{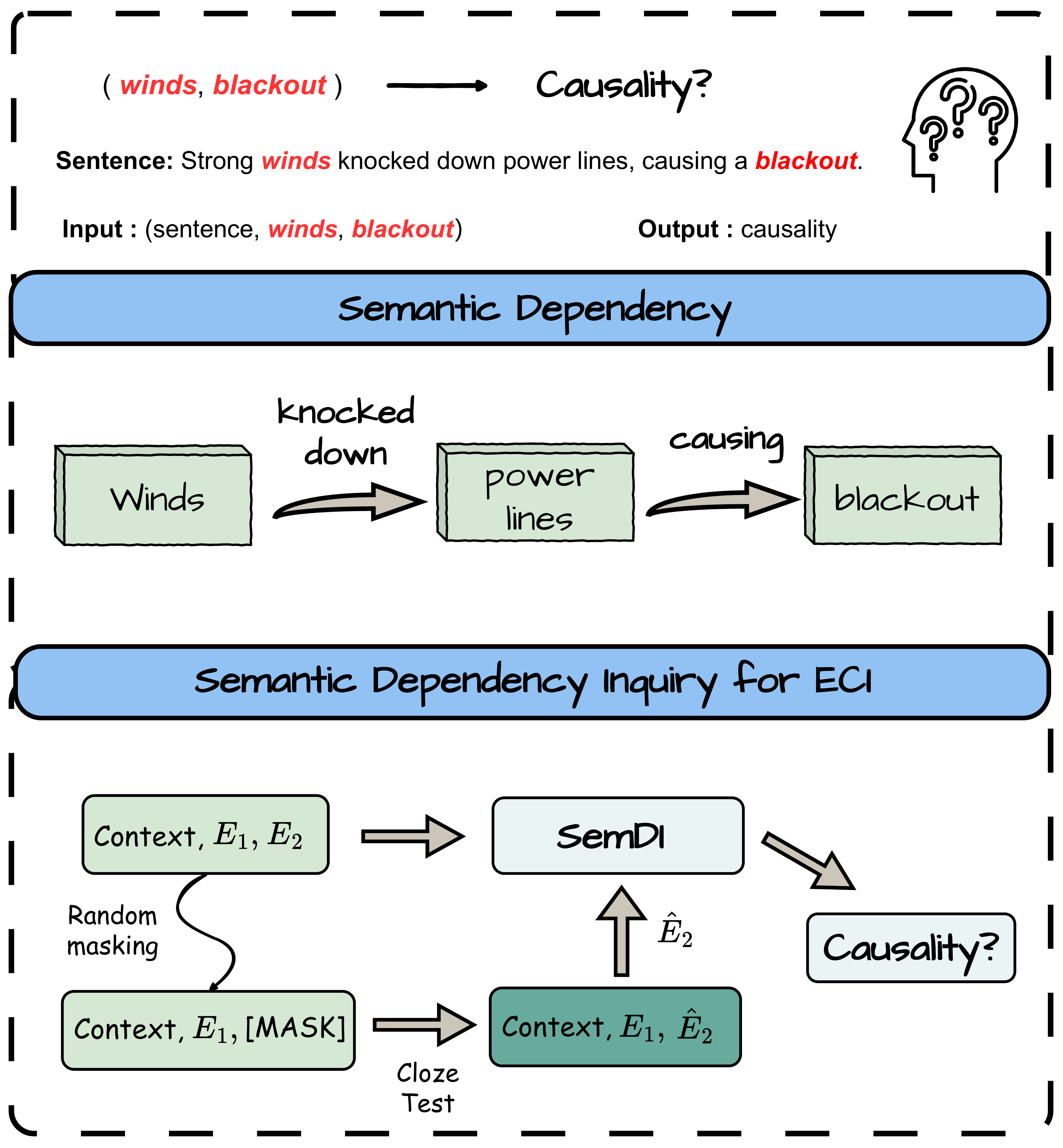}
    \caption{Introduction of the ECI task, along with our motivation: causal relations are heavily context-dependent.}
    \label{fig:intro}
    \vspace{-0.4cm}
\end{figure}

The conventional approach for ECI involves a binary classification model that takes a triplet (sentence, event-1, event-2) as input to determine the existence of a causal relation between the two events, as illustrated at the top of Figure~\ref{fig:intro}. Various methods have been proposed to enhance ECI performance. While early feature-based methods \citep{hashimoto-etal-2014-toward, ning-etal-2018-joint, gao-etal-2019-modeling} laid the foundation, more recent representation-based methods have demonstrated superior ECI capabilities, including Pre-trained Language Models (PLMs) based methods \citep{shen-etal-2022-event,man-etal-2022-event}, and data augmentation methods \citep{zuo-etal-2020-knowdis, zuo-etal-2021-learnda}. A notable recent trend is augmenting ECI models with external prior knowledge \citep{liu2021knowledge, cao2021knowledge, LIU2023110064}. However, it can also introduce potential bias. For example, consider the event pairs {\it \textbf{(winds, blackout)}} mentioned in Figure~\ref{fig:intro}. While there seems to be no direct causal relation from prior knowledge, contextual inference makes it reasonable to deduce causality. Upon analysis, we can observe a causal semantic dependency between "winds" and "blackout": {\it \textbf{winds}} $\xrightarrow{\text{knocked down}}$ {\it \textbf{power lines}} $\xrightarrow{\text{causing}}$ {\it \textbf{blackout}}. This reveals that causal relations between events within a sentence often appear as context-dependent semantic dependencies. Thus, we claim that the ECI task can be reformulated as a semantic dependency inquiry task between two events within the context. 

To this end, we propose a Heuristic Semantic Dependency Inquiry Network~(SemDI) for the ECI task. The key idea behind SemDI is to explore implicit causal relationships guided by contextual semantic dependency analysis. Specifically, we first capture the semantic dependencies using a unified encoder. 
Then, we randomly mask out one event from the event pair and utilize a {\it Cloze} analyzer to generate a fill-in token based on comprehensive context understanding. Finally, this fill-in token is used to inquire about the causal relation between the two events in the given sentence. The main contributions of this work are summarized as follows:

\begin{itemize}
  \item We propose the Semantic Dependency Inquiry as a promising alternative solution to the ECI task, highlighting the significance of contextual semantic dependency analysis in detecting causal relations.    
  \item We introduce a heuristic Semantic Dependency Inquiry Network~(SemDI) for ECI, which offers simplicity, effectiveness, and robustness.
  \item The experimental results on three widely used datasets demonstrate that SemDI achieves $7.1\%$, $10.9\%$, and $14.9\%$ improvements in F1-score compared to the previous SOTA methods, confirming its effectiveness.
\end{itemize}

%% file: 2-Related_work.tex
Identifying causal relationships between events in the text is challenging and has attracted massive attention in the past few years \citep{feder-etal-2022-causal}. Early approaches primarily rely on explicit causal patterns \citep{hashimoto-etal-2014-toward,riaz2014depth}, lexical and syntactic features \citep{riaz2013toward, riaz2014recognizing}, and causal indicators or signals \citep{do2011minimally, hidey2016identifying} to identify causality. 

Recently, representation-based methods leveraging Pre-trained Language Models (PLMs) have significantly enhanced the ECI performance. To mitigate the issue of limited training data for ECI, \citet{zuo-etal-2020-knowdis, zuo-etal-2021-learnda} proposed data augmentation methods that generate additional training data, thereby reducing overfitting. Recognizing the importance of commonsense causal relations for ECI, \citet{liu2021knowledge, cao2021knowledge, LIU2023110064} incorporated external knowledge from the knowledge graph ConceptNet \citep{speer2017conceptnet} to enrich the representations derived from PLMs. However, the effectiveness of external knowledge-based methods is highly contingent on the consistency between the target task domain and the utilized knowledge bases, which can introduce bias and create vulnerabilities in these approaches.

In contrast to previous methods, \citet{man-etal-2022-event} introduced a dependency path generation approach for ECI, explicitly enhancing the causal reasoning process. \citet{hu-etal-2023-semantic} exploited two types of
semantic structures, namely event-centered structure and event-associated structure, to capture associations between event pairs.





%% file: 3-Preliminaries.tex
\subsection{Problem Statement}
Let $\boldsymbol{S}=[S_1,\cdots, S_n] \in \mathbb{R}^{1 \times |S|}$ refer to a sentence with $|S|$ tokens, where each token $S_i$ is a word/symbol, including special identifiers to indicate event pair ($S_{e_1}$, $S_{e_2}$) in causality. Traditional ECI models determine if there exists a causal relation between two events by focusing on event correlations, which can be written as $\mathcal{F}(\boldsymbol{S}, S_{e_1}, S_{e_2})=\{0, 1\}$. Actually, correlation does not necessarily imply causation, but it can often be suggestive. Therefore, this study investigates the Semantic Dependency Inquiry (SemDI) as a potential alternative solution to the ECI task. For clarity, we introduce two fundamental problems:

\textbf{\textit{Cloze} Test.} We denote a mask indicator as $m=[m_1, \cdots, m_{|S|}\} \in \{ 0, 1 \}^{1 \times |S|}$, where $m_i=0$ if $S_i$ is event token, otherwise $m_j=1, j \in [1,\cdots,|S|], j \neq i$. We use $\boldsymbol{\hat{S}}$ instead of $\boldsymbol{S}$ to explicitly represent the incomplete sentence, i.e, $\boldsymbol{\hat{S}}=m\boldsymbol{S}$. For simplicity, if the event contains more than one word, we replace all words in the event with one '<MASK>' token. The {\it Cloze} test in this study is to develop a contextual semantic-based network $\boldsymbol{\Omega(\cdot)}$ to fill in the masked word, i.e., $\boldsymbol{\Omega(\boldsymbol{\hat{S}}}) \mapsto S_m$, where $S_m$ denotes the generated fill-in token.

\textbf{Semantic Dependency Inquiry.} There often exists a semantic dependency between two causally related events, as illustrated in Figure~\ref{fig:intro}. In light of this, we propose to inquire about such causal semantic dependency between two events within the context through the generated fill-in token. This approach aligns with our motivation that causal relations are heavily context-dependent. To elaborate,
given the input tuple ($\boldsymbol{S}, S_m$), a discriminator $\boldsymbol{\mathcal{D}(\cdot)}$ aims to examine the presence of causal semantic dependency
in sentence $\boldsymbol{S}$ through $S_m$, i.e., $\boldsymbol{\mathcal{D}(\boldsymbol{S}, S_m)} \in \{ 0,1 \}$.


\subsection{Basic Technique}
The multi-head attention mechanism is the core part of Transformer~\cite{vaswani2017attention} and has been widely adopted for sequential knowledge modeling. It measures the similarity scores between a given query and a key, whereafter formulating the attentive weight for a value. The canonical formulation can be conducted by the scaled dot-product as follows:
\begin{equation}
    \small
    \begin{aligned}
    \text{MHA}(A, B) &= \text{Concat}(H^1, \cdots, H^h), \\
    \text{where} \ H^i &= softmax(\frac{QK^T}{\sqrt{d_h}})V, \\
    \text{and} \ Q&=AW_Q, \{K, V\}=B\{W_K, W_V\},
    \end{aligned}
\end{equation}
herein, $W_{\{Q, K, V\}} \in \mathbb{R}^{d \times d_h}$ are head mapping parameters. Typically, the multi-head attention mechanism can be categorized into two types: (1) when $A=B$, the attention mechanism focuses on the relationship between different elements within the same input; (2) when $A \neq B$, the attention mechanism captures the relationship between elements from different inputs.

%% file: 4-Method.tex
\begin{figure*}
    \centering
    \includegraphics[width=1.\textwidth]
    {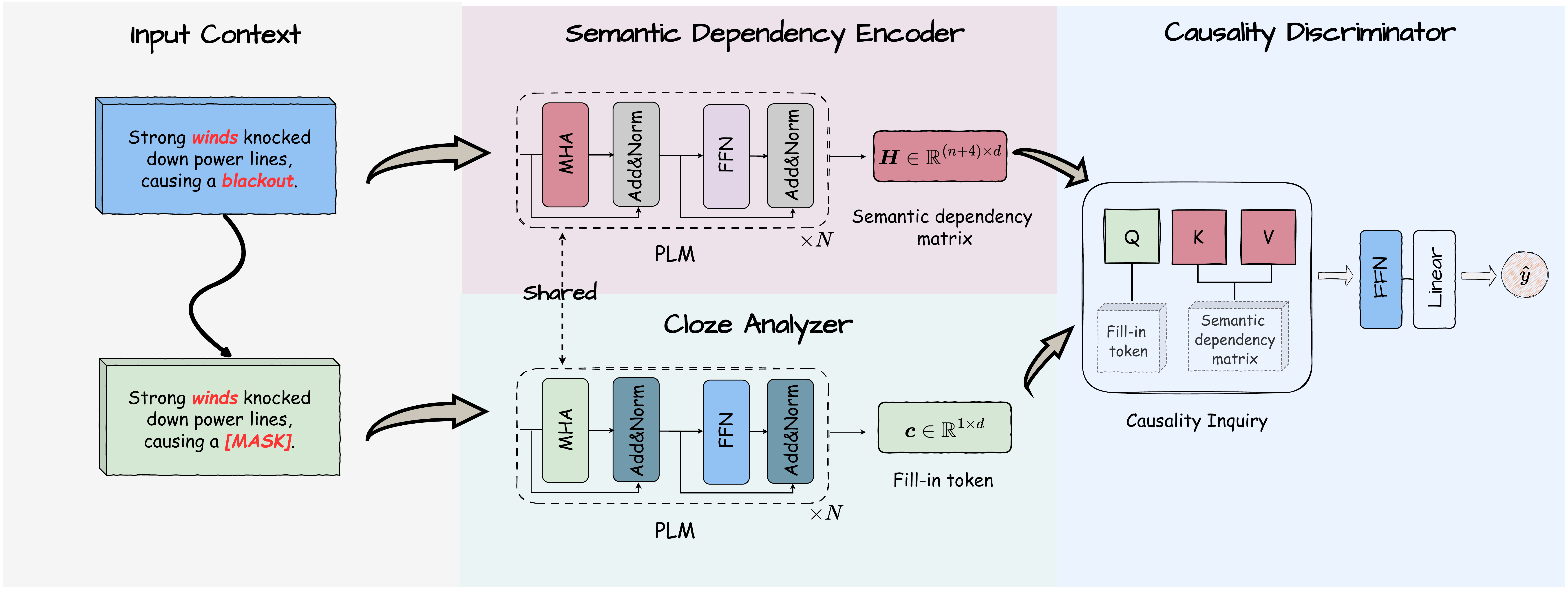} 
    \caption{Overview of our proposed SemDI for event causality identification, which consists of (1) a Semantic Dependency Encoder to capture the intricate semantic dependencies within the context; (2) a Cloze Analyzer to generate a fill-in token; (3) a Causality Discriminator to conduct causality inquiry.}
    \label{fig:model}
\end{figure*}



\subsection{Overview}
This section presents our proposed SemDI model, which reformulates the ECI task as a causal semantic dependency inquiry problem. As illustrated in Figure~\ref{fig:model}, we first capture the semantic dependencies within the source sentence using a Semantic Dependency Encoder (SDE). Then, we randomly mask out one event from the event pair and utilize a \textit{Cloze} Analyzer (CA) to generate a fill-in token based on comprehensive context understanding. Finally, this fill-in token is used to inquire about the causal semantic dependency between the two events in a Causality Discriminator. It is worth noting that the SDE and CA share the same parameters initialized from a Pre-trained Language Model (PLM), e.g., RoBERTa. The key distinguishing feature of our approach is its full utilization of reading comprehension within the generative model, eliminating the need for additional prior knowledge and prioritizing simplicity and efficiency.

\subsection{{\it Cloze} Analyzer}
\label{sec:ca}

It is reasonable to believe that a well-trained deep generative model is powerful in context awareness~\citep{goswami-etal-2020-unsupervised-relation}. In light of this, we adopt a straightforward approach of randomly masking one event from the event pair, and then predicting this event. This approach is inspired by the literary puzzle {\it Cloze}, which plays a crucial role in our framework. The {\it Cloze} facilitates the prediction of the most appropriate fill-in token for the masked word, thereby revealing the probable semantic relationships within the given context.

{\bf Input Embedding Layer} aims to encode sentences into a latent space. Given a sentence $\boldsymbol{S}=[S_1, \cdots, S_{e_1}, \cdots, S_{e_2}, \cdots, S_n]$, we correlate a $\boldsymbol{\hat{S}}=\boldsymbol{S} \odot \boldsymbol{M}_{mask}$, where $\odot$ denotes the element-wise product and $\boldsymbol{M}_{mask}=\{m_{1:n} \} \in \{0,1\}^n$ indicates the randomly masked word. If $m_i=0$, it means the $S_i$ word is masked, which can be either $S_{e_1}$ or $S_{e_2}$. In order to adhere to the {\it Cloze} puzzle setting, we utilize two pairs of specification symbols <$e_{1}$>, <$/e_{1}$> and <$e_{2}$>, <$/e_{2}$> to mark $S_{e_1}$ and $S_{e_2}$ in source sentence $\boldsymbol{S}$. Importantly, the masked word does not have the marker, thus resulting in $|\boldsymbol{\hat{S}}| = |\boldsymbol{S}| - 2$.

The input embedding layer encodes the $\boldsymbol{S}, \boldsymbol{\hat{S}}$ associated with its position. The word embeddings are trained along with the model and initialized from pre-trained RoBERTa word vectors with a dimensionality of $d=1024$. The specification symbol <$e_*$> and $[mask]$ are mapped to the appointed tokens, and their embeddings are trainable with random initialization. The position embedding is computed by the $sine$ and $cosine$ functions proposed by Transformer. Finally, the outputs of a given sentence from this layer are the sum of the word embedding and position embedding, namely $X$ and $\hat{X}$ for simplicity, respectively. The latter corresponds to a sentence with the masked word. Notably, $X \in {\mathbb{R}^{(n+4) \times d}, \hat{X} \in \mathbb{R}^{(n+2) \times d}}$.

{\bf Semantic Completion Block} receives the incomplete sentence $\hat{X}$ as input, aiming to fill in the blank that is marked by $[mask]$ (i.e., $\hat{x}_m$). We leverage a PLM, specifically RoBERTa, to address this \textit{Cloze} test. The main idea of this block is to take advantage of the $\hat{x}_m$ as a query, then fill the man-made gap. The process can be formulated as:

\begin{equation}
    \small
    \boldsymbol{c} = PLM(\hat{x}_m, \hat{X}), 
\end{equation}


where $\boldsymbol{c} \in \mathbb{R}^{1 \times d}$ is the output of this block, i.e., the embedding of the generated fill-in token.

\subsection{Semantic Dependency Encoder}
To capture the semantic dependencies between words within the context, we utilize a PLM, e.g., RoBERTa, as the Semantic Dependency Encoder to facilitate comprehensive information reception. It receives the source sentence $X$ as input to establish the semantic dependencies present in the entire sentence, which can be formulated as:
\begin{equation}
    \small
    \boldsymbol{H} = PLM (X),
\end{equation}
where $\boldsymbol{H} \in \mathbb{R}^{(n+4) \times d}$ denotes sentence representation that assimilate intricate semantic dependencies among words.


\subsection{Causality Discriminator}
According to our motivation, we conduct a causality inquiry between the fill-in token $\boldsymbol{c}$ and the semantic dependency matrix $\boldsymbol{H}$ by utilizing cross attentive network, namely:
\begin{equation}
    \small
    \boldsymbol{z} = \text{MHA}(\boldsymbol{c}, \boldsymbol{H}).
    \label{eq:CI-atten}
\end{equation}
After that, we obtain the $\boldsymbol{z} \in \mathbb{R}^{1 \times d}$ as the result of the inquiry. A two-layer feed-forward network transforms it to the causality classifier as:
\begin{equation}
    \small
    \boldsymbol{y}_z = (ReLU(\boldsymbol{z}W_{in} + b_{in})W_{out}+b_{out}),
\end{equation}
where $\{W_{*}, b_{*}\}$ are learnable parameters.

\subsection{Training Criterion}
We adopt the cross-entropy loss function to train SemDI:
\begin{equation}
    \scriptsize
    J(\Theta) = -\!\sum_{(s_{e_1}, s_{e_2}) \in \boldsymbol{S}} \!\boldsymbol{y}_{(s_{e_1}, s_{e_2})} \log \big( softmax(\boldsymbol{y}_zW_y + b_y) \big),
\end{equation}
where $\Theta$ denotes the model parameters, $\boldsymbol{S}$ refers to all sentences in the training set, $(s_{e_1}, s_{e_2})$ are the events pairs and $\boldsymbol{y}_{(s_{e_1}, s_{e_2})}$ is a one-hot vector indicating the gold relationship between $s_{e_1}$ and $s_{e_2}$. 
We utilize $\boldsymbol{y}_{(s_{e_1}, s_{e_2})}$ to guide the learning process in which the generated fill-in token is used to inquire about the causal semantic dependencies within the original sentence, as shown in Figure~\ref{fig:heatmap}.

It is worth noting that we do not establish a loss function to directly guide the generation of fill-in tokens. This decision is because we do not require alignment between the fill-in tokens and the original words. Instead, our objective is to generate a token based on comprehensive context understanding, which we then use to inquire about the presence of a causal relationship. This approach aligns with our main argument: {\it the existence of a causal relationship between two events is heavily context-dependent}.

%% file: 5-Experiments.tex
In this section, we empirically investigate the effectiveness of SemDI, aiming to answer the following questions: (1) Can SemDI consistently perform well across various ECI benchmarks? (2) Can the proposed moduls (e.g., \textit{Cloze} Analyzer) effectively enhance performance? (3) Does SemDI exhibit interpretability during the causality inquiry process? (4) Is SemDI robust to diffrent backbone sizes and masking strategies?

\subsection{Experimental Setup}
\label{sec:exp-setup}

\textbf{Evaluation Benchmarks.} We evaluate our SemDI on three widely-used ECI benchmarks, including two from EventStoryLine v0.9 \citep{caselli-vossen-2017-event} and one from Causal-TimeBank \citep{mirza-etal-2014-annotating}, namely ESC, ESC\textsuperscript{*}, and CTB. \textbf{ESC}\footnote{\url{https://github.com/tommasoc80/EventStoryLine}} contains $22$ topics, $258$ documents, and $5334$ event mentions. This corpus contains $7805$ intra-sentence event pairs, of which $1770$ ($22.67\%$) are annotated with causal relations. \textbf{ESC\textsuperscript{*}} is a different partition setting of the ESC dataset, utilized by \citet{man-etal-2022-event, shen-etal-2022-event, hu-etal-2023-semantic}. Unlike the original ESC dataset, which sorts documents by topic IDs, this setting involves random shuffling of documents, leading to more consistent training and testing distributions. \textbf{CTB} \footnote{\url{https://github.com/paramitamirza/Causal-TimeBank}} consists of $183$ documents and 6811 event mentions. Among the $9721$ intra-sentence event pairs, $298$ ($3.1\%$) are annotated with causal relations. Table~\ref{tab:dataset-stat} provides statistics of these benchmarks. More detailed descriptions are discussed in Appendix~\ref{app:dataset}.

\begin{table}[h]
\centering
\small
\caption{Statistics of evaluation benchmarks, where OOD denotes Out-of-Distribution, ID denotes In-Distribution, and CI denotes Class Imbalance.}
\label{tab:dataset-stat}
\begin{tabularx}{\columnwidth}{lXXXX}
\toprule
\textbf{Dataset} & \textbf{\# Doc} & \textbf{\# Pairs} & \textbf{\# Causal} & \textbf{Evaluation} \\ 
\midrule
ESC       & 258              & 7805           & 1770            & OOD \\
ESC*      & 258         & 7805           & 1770            & ID \\
CTB       & 183        & 9721           & 298             & CI\\ 
\bottomrule
\end{tabularx}
\end{table}

\textbf{Baselines.} We first compare our proposed SemDI with the feature-based methods. For the ESC dataset, we adopted the following baselines: \textbf{LSTM} \citep{cheng-miyao-2017-classifying}, a dependency path boosted sequential model; \textbf{Seq} \citep{choubey-huang-2017-sequential}, a sequence model explores manually designed features for ECI. \textbf{LR+} and \textbf{ILP} \citep{gao-etal-2019-modeling}, models considering document-level structure. For the CTB dataset, we select \textbf{RB} \citep{mirza2014analysis}, a rule-based ECI system; \textbf{DD} \citep{mirza2016catena}, a data-driven machine learning-based method; \textbf{VR-C} \citep{mirza2014extracting}, a verb rule-based model boosted by filtered data and causal signals.

Furthermore, we compare SemDI with the following PLMs-based methods: \textbf{MM} \citep{liu2021knowledge}, a commonsense knowledge enhanced method with mention masking generalization; \textbf{KnowDis} \citep{zuo-etal-2020-knowdis}, a knowledge-enhanced distant data augmentation approach; \textbf{LearnDA} \citep{zuo-etal-2021-learnda}, a learnable augmentation framework alleviating lack of training data; \textbf{LSIN} \citep{cao2021knowledge}, an approach which constructs a descriptive graph to exploit external knowledge; \textbf{CauSeRL} \citep{zuo-etal-2021-improving}, a self-supervised method utilizing external causal statements; \textbf{GenECI} and \textbf{T5 Classify}~\cite{man-etal-2022-event}, methods that formulates ECI as a generation problem; \textbf{KEPT}~\cite{LIU2023110064}, a study that leverages BERT to integrate external knowledge bases for ECI; \textbf{SemSIn}~\citep{hu-etal-2023-semantic}, the previous SOTA method that leverages event-centric structure and event-associated structure for causal reasoning. Similar to our approach, it does not utilize external knowledge; 

We also compare SemDI with other state-of-the-art Large Language Models (LLMs), including \textbf{GPT-3.5-turbo}, \textbf{GPT-4} \citep{achiam2023gpt}, and \textbf{LLaMA2-7B} \citep{touvron2023llama}. These models are known for their extensive pre-training on diverse datasets and their superior performance across multiple tasks.

\textbf{Implementation Details.} We adopt the commonly used {\bf P}recision, {\bf R}ecall, and {\bf F1}-score as evaluation metrics. Following the existing studies \citep{shen-etal-2022-event, hu-etal-2023-semantic, LIU2023110064}, we select the last two topics in ESC as development set and use the 
remaining $20$ topics for a $5$-fold cross-validation. In addition, we perform a $10$-fold cross-validation on CTB. Given the sparsity of causality in the CTB dataset, we follow \citet{cao2021knowledge, hu-etal-2023-semantic} to conduct a negative sampling technique for training with a sampling rate of $0.7$. The pre-trained RoBERTa-large model \citep{liu2019roberta} is chosen as the backbone of our Cloze Analyzer and Semantic Dependency Encoder. The hidden dimension is $1024$, the batch size is $20$, and the dropout rate is $0.5$. We train our model via the AdamW~\cite{loshchilov2017decoupled} optimizer with an initial learning rate of $1e-5$. The entire training process spans $100$ epochs and takes approximately $2$ hours. Additionally, we fine-tune the Llama-2-7b-chat-hf \citep{touvron2023llama} using the LlamaFactory \citep{zheng2024llamafactory}. Detailed prompts guiding LLMs to identify causality are provided in Appendix~\ref{app:prompt}. All experiments are conducted on one Nvidia GeForce RTX $3090$.

\subsection{Main Results}


\begin{table}[ht]
    \centering
    \resizebox{\columnwidth}{!}{
        \begin{tabular}{l|c|c|c}
            \toprule
              \textbf{Method}& \textbf{P} & \textbf{R}& \textbf{F1} \\
             \midrule
             LSTM \citep{cheng-miyao-2017-classifying} & 34.0 & 41.5 & 37.4\\ 
             Seq \citep{choubey-huang-2017-sequential} & 32.7 & 44.9 & 37.8\\
             LR+ \citep{gao-etal-2019-modeling} & 37.0 & 45.2 & 40.7\\
             ILP \citep{gao-etal-2019-modeling} & 37.4 & 55.8 & 44.7\\
             KnowDis \citep{zuo-etal-2020-knowdis} & 39.7 & 66.5 & 49.7\\
             MM \citep{liu2021knowledge} & 41.9 & 62.5 & 50.1\\
             CauSeRL \citep{zuo-etal-2021-improving} & 41.9 & 69.0 & 52.1\\
             LSIN \citep{cao2021knowledge} & 49.7 & 58.1 & 52.5\\
             LearnDA \citep{zuo-etal-2021-learnda} & 42.2 & \underline{69.8} & 52.6\\
             SemSIn \citep{hu-etal-2023-semantic} & \underline{50.5} & 63.0 & 56.1 \\
             KEPT \cite{LIU2023110064} & 50.0 & 68.8 & 
             \underline{57.9} \\
             \midrule
              LLaMA2-7B
              & 11.4 & 50.0 & 18.6 \\
              LLaMA2-7B\textsuperscript{\textit{ft}} 
              & 20.5 & 57.1 & 29.8 \\
              GPT-3.5-turbo  
              & 39.5 & 40.3 & 39.7\\
              GPT-4.0 
              & 30.7 & {\bf 85.7} & 45.2\\
              ${\bf SemDI}$ & {\bf 56.7} & {68.6} & {\bf 62.0} \\ 
             
             \midrule
             \midrule
             T5 Classify\textsuperscript{*} \citep{man-etal-2022-event} & 39.1 & 69.5 & 47.7\\
             GenECI\textsuperscript{*} \citep{man-etal-2022-event} & 59.5 & 57.1 & 58.8\\ 
             SemSIn\textsuperscript{*} \citep{hu-etal-2023-semantic} & 64.2 & 65.7 & 64.9\\
             DPJL\textsuperscript{*} \citep{shen-etal-2022-event} & 
             \underline{65.3} & 70.8 & \underline{67.9}\\
             \midrule
             LLaMA2-7B
             & 12.1 & 50.7 & 19.5 \\
             LLaMA2-7B\textsuperscript{\textit{ft}*} 
             & 20.3 & 57.6 & 30.0 \\
             GPT-3.5-turbo\textsuperscript{*} 
             & 40.1 & 41.2 & 40.6 \\
             GPT-4.0\textsuperscript{*} 
             & 31.2 & {\bf 86.3} & 45.8\\             
            ${\bf SemDI}^{*}$ & {\bf 75.0} & \underline{75.7} & {\bf 75.3} \\ 
             \bottomrule
        \end{tabular}
    }
     \caption{Experimental results on ESC and ESC\textsuperscript{*}. * denotes experimental results on ESC\textsuperscript{*} and \textit{ft} denotes fine-tuning the LLM.}
     \label{tab:res-esc}
\end{table}

\begin{table}[ht]
    \centering
    \resizebox{\columnwidth}{!}{
        \begin{tabular}{l|c|c|c}
            \toprule
              \textbf{Method}& \textbf{P} & \textbf{R}& \textbf{F1} \\
             \midrule
             RB \citep{mirza2014analysis} & 36.8 & 12.3 & 18.4\\ 
             DD \citep{mirza2016catena} & \underline{67.3} & 22.6 & 33.9\\
             VR-C\citep{mirza2014extracting}  & \bf 69.0 & 31.5 & 43.2\\
             MM \citep{liu2021knowledge} & 36.6 & 55.6 & 44.1\\
             KnowDis \citep{zuo-etal-2020-knowdis} & 42.3 &60.5 &49.8\\
             LearnDA \citep{zuo-etal-2021-learnda} &41.9 & 68.0 & 51.9\\
             LSIN  \citep{cao2021knowledge} &51.5 &56.2 &52.9\\
             CauSeRL \citep{zuo-etal-2021-improving} &43.6 & 68.1 & 53.2\\
             KEPT \citep{LIU2023110064} & 48.2 & 60.0 & 53.5\\
             GenECI  \citep{man-etal-2022-event} & 60.1 & 53.3 &56.5\\
             SemSIn \citep{hu-etal-2023-semantic} & 52.3 & 65.8 & \underline{58.3} \\
             \midrule
             LLaMA2-7B 
             & 5.4 & 53.9 & 9.8 \\
             LLaMA2-7B\textsuperscript{\textit{ft}} 
             
             & 10.5 & 61.8 & 17.9 \\
             GPT-3.5-turbo
             & 7.0 & 49.7 & 12.3 \\
             GPT-4.0 
             & 4.6 & {\bf 84.6} & 8.7 \\
             ${\bf SemDI}$ & 59.3 & \underline{77.8} & {\bf 67.0} \\ 
             \bottomrule
        \end{tabular}
    }
     \caption{Experimental results on CTB. \textit{ft} denotes fine-tuning the LLM.}
     \vspace{-0.3cm}
     \label{tab:res-ctb}
\end{table}

Table~\ref{tab:res-esc} and Table~\ref{tab:res-ctb} present the performance of different approaches on three benchmarks, respectively. 
The best scores are highlighted in {\bf bold}, while the second-best scores are \underline{underlined}. We summarize our observations as follows:

\textbf{SemDI consistently outperforms all baselines in terms of the F1-score.} More specifically, SemDI surpasses the previous SOTA methods by significant margins of $4.1, 7.4$, and $8.7$ in F1-score on the ESC, ESC\textsuperscript{*}, and CTB datasets, respectively. This result aligns with our motivation, as prioritizing the context-dependent nature of causal relations enables the model to identify causality more accurately, thereby mitigating potential bias introduced by external prior knowledge.

\textbf{Domain Generalization Ability.} On the ESC dataset, ECI models need to generalize to test topics $\mathcal{D}_{test}$ that are disjoint from the training topics $\mathcal{D}_{train}$, i.e., $\mathcal{D}_{train} \cap \mathcal{D}_{test} = \varnothing$. From Table~\ref{tab:res-esc}, we observe that SemDI demonstrates superior performance under this Out-of-Distribution (OOD) testing. This result verifies SemDI's potential as a general framework for event causality identification. Furthermore, training and testing distributions are more consistent under the ESC\textsuperscript{*} dataset, resulting in relatively higher performance.

{
\renewcommand{\arraystretch}{1.15}
\begin{table*}[ht]
\centering
\begin{tabular}{c|ccc|ccc|ccc}
\toprule[1pt]
\multicolumn{1}{c|}{\multirow{2}{*}{\textbf{Method}}} & \multicolumn{3}{c|}{\textbf{ESC}} & \multicolumn{3}{c|}{\textbf{ESC\textsuperscript{*}}} & \multicolumn{3}{c}{\textbf{CTB}} \\
\multicolumn{1}{c|}{}                        & P      & R      & F1     & P      & R     & F1    & P      & R     & F1 \\ 
\midrule[1pt]
SemDI w/o. CA                                      & {\bf 57.8} & 64.0 & 60.8        &  74.8 &  75.2 & 74.9  &   63.8   &   65.0  &   63.9 \\
SemDI w/o. SDE                                     & 56.8 & 57.9 & 56.9  &  67.2      &  68.9  &  68.0  &  {\bf 64.4}   &   61.9 & 62.5   \\
SemDI w/o. RoBERTa                                & 52.2 & 68.5 &   59.1 &  70.9  &  73.0   &  71.9  &  59.1  &  66.4  &  61.0\\
\midrule[0.5pt]
\textbf{SemDI}               & 56.7 & {\bf 68.6} & {\bf 62.0}  &   {\bf 75.0}     &    {\bf 75.7}   & {\bf 75.3} &  59.3  &   {\bf 77.8}   &  {\bf 67.9} \\   
\bottomrule[1pt]
\end{tabular}
\caption{Results of ablation study, which demonstrates the impact of different components on the overall performance of our model.}
\label{tb:ablation}
\end{table*}
}


\textbf{Comparison with PLMs-based Methods.} Compared to LearnDA, which achieves the second-highest Recall score on the ESC dataset (at the top of Table~\ref{tab:res-esc}), SemDI shows a significant improvement of $34.3\%$ in Precision. This indicates that SemDI is more reliable in decision-making. It is understandable that LearnDA achieves better recall, as it can generate additional training event pairs beyond the training set. While KEPT shares the same fundamental architecture with SemDI, it mainly focuses on integrating external knowledge for causal reasoning. In contrast, SemDI highlights the importance of contextual semantic dependency analysis, outperforming KEPT by a significant margin.

\textbf{Comparison with LLMs.} Our SemDI model demonstrates superior performance compared to state-of-the-art Large Language Models (LLMs) across all benchmarks, despite its significantly smaller size. Specifically, SemDI ($368$M parameters) is $19$ times smaller than fine-tuned LLaMA2-7B, yet it achieves an average improvement of $177.8$\% in F1-score. The efficiency of SemDI makes it ideal for deployment in resource-constrained and time-demanding environments. Additionally, we observe that LLMs often exhibit overconfidence in determining causal relationships, resulting in high recall but low precision. This observation is consistent with previous findings in the literature \citep{si-etal-2022-examining, mielke-etal-2022-reducing, xiong2024llms}.

\subsection{Ablation Study}

In this subsection, we conduct comprehensive ablation studies to demonstrate the effectiveness of our key components, including 
the \textit{Cloze} Analyzer (CA), the Semantic Dependency Encoder (SDE), and the backbone model RoBERTa. Concretely, we remove \textit{Cloze} Analyzer and utilize the original event embedding for causality inquiry in {\bf SemDI w/o CA}. In {\bf SemDI w/o SDE}, we remove the Semantic Dependency Encoder and directly feed the 
embedding of the generated fill-in token to the classifier, thus omitting the causality inquiry process. In {\bf SemDI w/o RoBERTa}, we replace the backbone RoBERTa-large model with a BERT-large model. The results are shown in Table~\ref{tb:ablation}.



From this table, we observe that: (1) SemDI outperforms all the variants, demonstrating the effectiveness of multiple components in SemDI, including the generation of fill-in token for causality inquiry, the encoding of semantic dependency, and the backbone selection. (2) SemDI w/o CA performs worse than SemDI, which indicates the importance of using a generated fill-in token to perform causality inquiry. Using the original token embedding that lacks the comprehensive context understanding for causality inquiry will lead to performance degradation. (3) SemDI w/o SDE shows the worst performance. This result is not surprising, as the analysis and inquiry of semantic dependency play the most crucial role in our approach to detecting causal relations. (4) Even if we replace the backbone RoBERTa model with a less optimized BERT model, our approach still outperforms the existing SOTA methods, including KEPT, SemSIn, and GPT-4.0, whose results are shown in Table~\ref{tab:res-esc} and Table~\ref{tab:res-ctb}. This further supports our claim that comprehensive contextual analysis is crucial for identifying causal relations within sentences.

\begin{figure}[h!]
    \vspace{0.4cm}
    \centering    
        \setlength{\abovecaptionskip}{1pt}
	\setlength{\belowcaptionskip}{1pt}
    \includegraphics[width=\columnwidth]{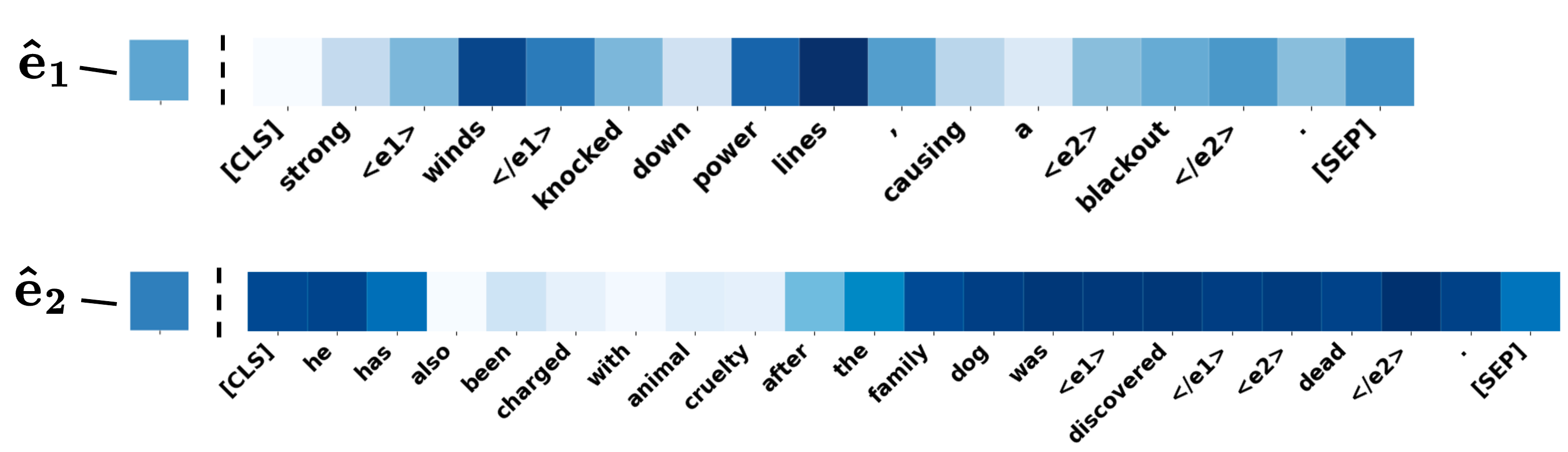} 
    \caption{Visualization of the attention heatmap in the causality inquiry process. Token "$\hat{e}_{*}$" denotes the generated fill-in token for event $e_{*}$.}
    \label{fig:heatmap}
    \vspace{-0.3cm}
\end{figure}

\subsection{Interpretability Analysis} 
\label{sec:interpre}

In this subsection, we visualize the causality inquiry process in SemDI to demonstrate its interpretability. Specifically, in this process, the generated fill-in token is used to inquire about the causal semantic dependencies between two events within the context, as shown in the middle of Figure~\ref{fig:intro}. We randomly select two examples from the ESC dataset and present their attention heatmap of the causality inquiry process in Figure~\ref{fig:heatmap}. It can be observed that the causality inquiry process can effectively uncover the intricate semantic dependencies between two events. For example, SemDI tends to uniformly distribute its attention to the sentence with non-causal event pairs, as shown in the heatmap of the second sentence. In contrast, we can observe a clear causal semantic dependency between "winds" and "blackout" in the heatmap of the first sentence: \textit{winds} $\rightarrow$ \textit{power lines} $\rightarrow$ \textit{blackout}. This phenomenon not only supports our motivation that causal relations are heavily context-dependent, but also demonstrates the effectiveness of using generated fill-in token to inquire about such causal semantic dependencies.


\begin{table*}[htbp!]
    \centering
    \resizebox{\textwidth}{!}{
    \begin{tabular}{m{10cm}|m{2cm}<{\centering}| m{2cm}<{\centering}| m{2cm}<{\centering}| m{2cm}<{\centering} } 
        \toprule       
        \textbf{Sentence} & \textbf{Masked} & \textbf{Fill-in} & \textbf{Golden} & \textbf{SemDI} \\
        \midrule
        A goth was being \textbf{questioned} on suspicion of \textbf{murder} yesterday after his mother and sister were found dead at home. & questioned & investigated & \Checkmark & \Checkmark \\
        \midrule
        A Kraft Foods plant worker who had been \textbf{suspended} for feuding with colleagues, then \textbf{escorted} from the building, returned minutes later with a handgun, found her foes in a break room and executed two of them with a single bullet each and critically wounded a third, police said Friday. & escorted   & retired  & \XSolidBrush & \Checkmark \\
        \bottomrule
    \end{tabular}
    }
    \caption{Case studies of SemDI. Two examples are randomly selected from the testing set of the ESC dataset.}
    \label{tab:case}
\end{table*}

\subsection{Robustness Analysis} \label{sec:sensi}
We now evaluate how different selections of key hyper-parameters impact our model's performance. 

\begin{figure}[h]
    \vspace{0.1cm}
    \centering
    \includegraphics[width=0.75\columnwidth, height=0.65\columnwidth]{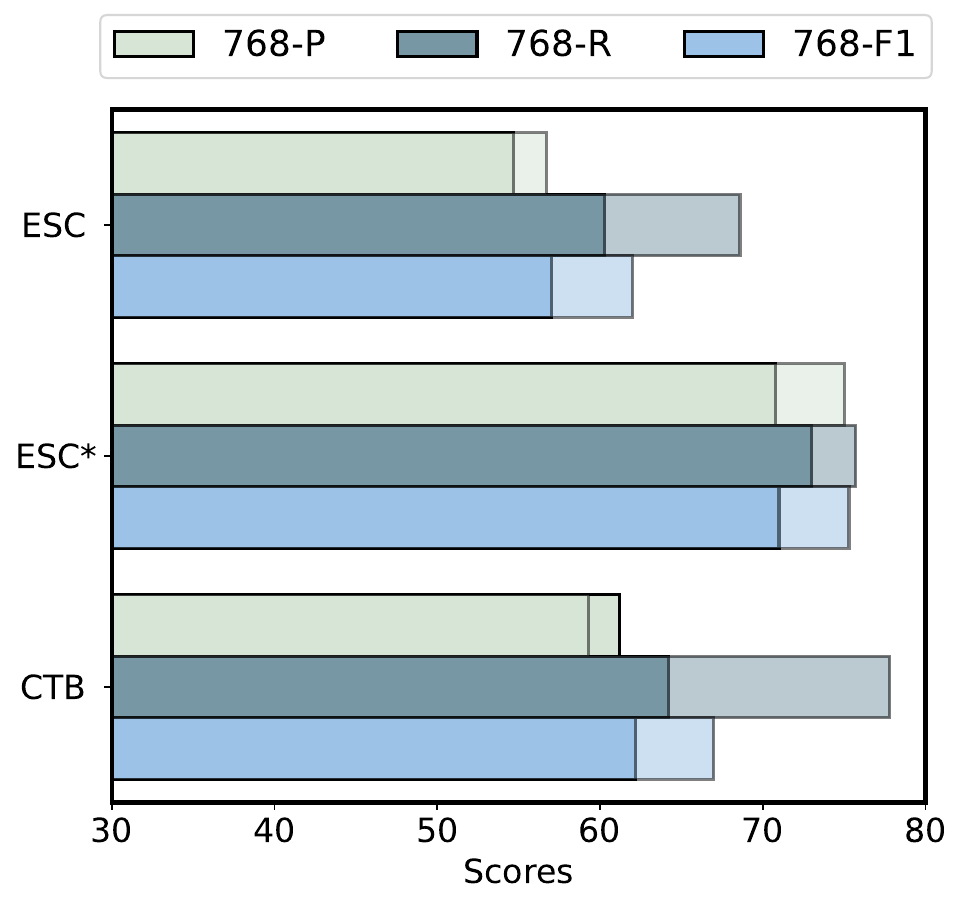}
    \caption{Robustness analysis on hidden size. The shaded portion represents hidden size $ = 1024$.}
    \label{fig:sensi-hidden}
    \vspace{0.2cm}
\end{figure}

{\bf Impact of hidden size.} We further analyze the impact of hidden size on two classic dimensions, $768$ and $1024$, as depicted in Figure~\ref{fig:sensi-hidden}, where the shaded portion corresponds to $1024$. From these results, we observe that: (1) Even if we reduce the hidden size from $1024$ to $768$, our SemDI still outperforms the previous SOTA methods, confirming its effectiveness and robustness. (2) The overall performance of SemDI shows a significant improvement with an increase in hidden size, particularly for the CTB dataset. This phenomenon can be attributed to the enhanced representation 
capability brought by higher model dimensions \citep{kaplan2020scaling}, which in turn facilitate reading comprehension - the core part of SemDI. (3) SemDI is relatively sensitive to the hidden size under low-resource scenarios (CTB) while maintaining good performance with sufficient annotated data for training (ESC and ESC\textsuperscript{*}).

{\bf Impact of masking strategy.} In Sec~\ref{sec:ca}, we randomly mask out one event from the event pair and then utilze a {\it Cloze} Analyzer to generate a fill-in token. To evaluate our model's sensitivity to the masking strategy applied in this {\it Cloze} test, we conduct further experiments on the ESC dataset with three specific approaches: (1) randomly mask $e_1$ or $e_2$ with a $50/50$ chance (Random); (2) "$100\%$ mask $e_1$" (Event1 only); (3) "$100\%$ mask $e_2$" (Event2 only). As shown in Table~\ref{tab:sensi-mask}, our SemDI maintains superior performance under all approaches in terms of the F1-score, confirming its robustness to varying masking strategies.


{
\renewcommand{\arraystretch}{1.2}
\begin{table}[]
\centering
\begin{tabular}{l|ccc}
\toprule
Strategy   & P    & R    & F1   \\ 
\midrule
Random      & 56.7 & 68.8 & 62.0 \\
Event1 only & {\bf 58.2} & 68.0 & {\bf 62.7} \\
Event2 only & 55.5 & {\bf 70.0} & 61.8 \\ 
\bottomrule
\end{tabular}
\caption{Robustness analysis on masking strategy applied in the \textit{Cloze} Test.}
\label{tab:sensi-mask}
\vspace{-0.1cm}
\end{table}
}

\vspace{0.2cm}
\subsection{Case Studies}
\label{app:Case Study}
In this subsection, we present case studies in Table~\ref{tab:case} to further analyze the performance of SemDI. It is worth noting that tied embeddings are employed to map the fill-in tokens to specific words. In case $1$, we can observe a clear causal semantic dependency: {\it \textbf{murder}} $\xrightarrow{\text{causing}}$ {\it \textbf{questioned}}. With a comprehensive understanding of the context, the \textit{Cloze} Analyzer can generate a fill-token that fits seamlessly within the given context, i.e., \textit{\textbf{(questioned, investigated)}}. Case $2$ demonstrates a faulty decision, likely due to the complex multi-hop reasoning required. Interestingly, the fill-in token "retired" also sharply contrasts with the original word "escorted." This misalignment may suggest a failure of SemDI to understand the semantic dependency between two events within the context.

%% file: 6-Appendix.tex
\subsection{Prompt}
\label{app:prompt}
In Sec~\ref{sec:exp-setup}, we utilize a prompt to guide the LLMs, including GPT-3.5-turbo, GPT-4, and LLaMA2-7B, to identify causal relations between two events within the sentence. We detail the prompt as follows.

"Given a sentence: \{sentence\}, decide if there exists a causal relation between \{event\_1\} and \{event\_2\} in this sentence. Your answer should be yes or no."

We also provide two examples from the ESC and CTB dataset in Table~\ref{tab:prompt}.

\begin{table}[h!] \normalsize
\centering
\begin{tabularx}{\columnwidth}{XX}
\toprule[1pt]
\textbf{ESC} \\ 
Given a sentence: "Strong winds knocked down power lines, causing a blackout.", decide if there exists a causal relation between "winds" and "blackout" in this sentence. Your answer should be yes or no.\\ 
\midrule [0.5pt]
\textbf{CTB}\\ 
Given a sentence: "He indicated that some assets might be sold off to service the debt.", decide if there exists a causal relation between "indicated" and "service" in this sentence. Your answer should be yes or no. \\
\bottomrule[1pt]
\end{tabularx}
\caption{Examples of prompt guiding LLMs to identify causal relations.}
\label{tab:prompt}
\end{table}

\subsection{Dataset Description}
\label{app:dataset}
In this subsection, we provide detailed descriptions for the three datasets we used in
experiments, i.e., ESC, ESC\textsuperscript{*}, and CTB.
\begin{itemize}
    \item \textbf{ESC.} This dataset contains 22 topics, 258 documents, and 5334 event mentions. The same as \citep{gao-etal-2019-modeling}, we exclude aspectual, causative, perception, and reporting event mentions, since most of which were not annotated with any causal relation. After the data processing, there are $7805$ intra-sentence event mention pairs in the corpus, $1770$ ($22.67\%$) of which are annotated with a causal relation. Identical to the data split in previous methods \citep{hu-etal-2023-semantic, zuo-etal-2021-learnda}, we select the last two topics in ESC as development set and use the remaining $20$ topics for a $5$-fold cross-validation. Note that the documents are sorted according to their topic IDs under this data partition setting, which means that the training and test sets are cross-topic. Due to the distribution gap between the training and test sets, the domain generalization ability of the model can be better evaluated.
    \item \textbf{ESC\textsuperscript{*}.} This dataset is a different partitioning of the ESC dataset. More specifically, it randomly shuffles the documents before training. Therefore, the distributions of the training and test sets are more consistent, because both two sets contain data on all topics. The experimental results under this setting can better demonstrate the model's ability to identify causal relations in topic-centered documents, which are common in real-world scenarios.
    \item \textbf{CTB.} CTB consists of $183$ documents and 6811 event mentions. Among the $9721$ intra-sentence event pairs, $298$ ($3.1\%$) are annotated with causal relations. Given the sparsity of causality in the CTB dataset, we follow existing works \citep{cao2021knowledge, hu-etal-2023-semantic} to conduct a negative sampling technique for training with the sampling rate of 0.7.
\end{itemize}